\documentclass[11pt]{article}

\usepackage[final]{acl}

\usepackage{times}
\usepackage{latexsym}
\usepackage[T1]{fontenc}
\usepackage[utf8]{inputenc}

\usepackage{microtype}

\usepackage{inconsolata}

\usepackage{graphicx}

\usepackage{tikz}
\usetikzlibrary{arrows.meta, decorations.pathreplacing}

\usepackage{hyperref}
\usepackage{url}
\usepackage{booktabs}
\usepackage{multirow}
\usepackage{comment}
\usepackage{tcolorbox}
\usepackage{tabularx}
\usepackage[table]{xcolor}
\usepackage{fontawesome5}
\usepackage{siunitx}
\usepackage{diagbox}
\usepackage{tcolorbox}
\tcbuselibrary{skins}
\usepackage{xcolor}
\usepackage{enumitem}
\usepackage{amssymb}
\usepackage{arydshln}
\usepackage{subcaption}
\usepackage{caption}
\usepackage{amsmath}
\usepackage{wrapfig}
\usepackage{tikz}
\usetikzlibrary{arrows.meta, positioning, shapes.geometric, shadows}

\usepackage{graphicx}
\newcommand{\hflogo}{\raisebox{-0.2\height}{\includegraphics[height=1em]{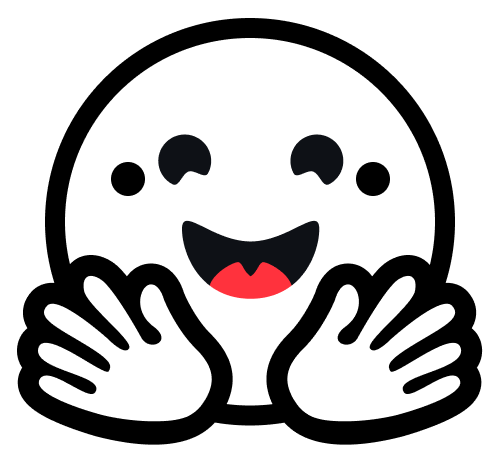}}}

\newcommand{\openthinker}{\raisebox{-0.2\height}{\includegraphics[height=1.4em]{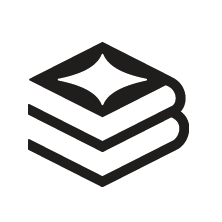}}}

\newcommand{\tmid}{{\footnotesize\texttt{|--}}\,}
\newcommand{\tend}{{\footnotesize\texttt{`--}}\,}

\definecolor{inputcolor}{RGB}{70,130,180}
\definecolor{reasoncolor}{RGB}{160,82,45}
\definecolor{outputcolor}{RGB}{60,120,60}
\definecolor{inputcolor}{RGB}{0,114,178}    % blue
\definecolor{reasoncolor}{RGB}{213,94,0}    % vermillion/orange
\definecolor{outputcolor}{RGB}{0,158,115}   % bluish green
\definecolor{rowgray}{gray}{0.95}
\definecolor{boxbg}{RGB}{235, 235, 245} % light lavender-gray
\definecolor{lightyellow}{RGB}{255, 249, 196}
\definecolor{lightgreen}{RGB}{220, 245, 220}
\definecolor{lightolivegreen-3}{RGB}{210, 225, 160}

\newcommand{\field}[2]{\textbf{\textcolor{#1}{#2}}}

\usepackage{lineno}

\definecolor{darkblue}{rgb}{0, 0, 0.5}
\hypersetup{colorlinks=true, citecolor=darkblue, linkcolor=darkblue, urlcolor=darkblue}

\newcommand{\reasonxlLogoD}{
\begin{tikzpicture}[baseline=-0.8ex, scale=0.18]
  \node[circle, draw, inner sep=1pt] (c) at (0,0) {};
  \node[circle, draw, inner sep=1pt] (a) at (1,0.8) {};
  \node[circle, draw, inner sep=1pt] (b) at (1,-0.8) {};
  \node[circle, draw, inner sep=1pt] (d) at (2,0) {};
  \draw[thick] (c)--(a)--(d)--(b)--(c);
\end{tikzpicture}
}

\newcommand{\reasonxl}{\textbf{\texttt{Reason}}\textsubscript{\textbf{XL}}}

\title{\reasonxlLogoD \reasonxl: A Large-Scale Corpus \\ for Language-Specific Reasoning in LLMs}

\author{Daniil Gurgurov\textsuperscript{*$\textcolor{blue}{D},\textcolor{green!50!black}{S}$} \quad Tom Röhr\textsuperscript{*$\textcolor{red}{B}$} \quad Sebastian von Rohrscheidt\textsuperscript{$\textcolor{red}{B}$}\\
\textbf{Josef van Genabith\textsuperscript{$\textcolor{blue}{D},\textcolor{green!50!black}{S}$}} \quad \textbf{Alexander Löser\textsuperscript{$\textcolor{red}{B}$}} \quad \textbf{Simon Ostermann\textsuperscript{$\textcolor{blue}{D},\textcolor{green!50!black}{S}$}} \\
\\
\textsuperscript{$\textcolor{blue}{D}$} German Research Center for Artificial Intelligence (DFKI)\\  \textsuperscript{$\textcolor{green!50!black}{S}$} Saarland University \quad \textsuperscript{$\textcolor{red}{B}$} Berliner Hochschule für Technik \\
\texttt{\small daniil.gurgurov@dfki.de, troehr@bht-berlin.de} \\
\textsuperscript{*} \scriptsize These authors contributed equally.
}

\begin{document}
\maketitle

\begin{abstract}
Despite advances in multilingual capabilities, most large language models (LLMs) remain English-centric in their training and, crucially, in the reasoning traces they produce: even for non-English problems, they predominantly reason in English, leaving non-English users with intermediate reasoning they cannot follow.
We address this disparity with two contributions.
\textcolor{inputcolor}{\textbf{(a)}} We introduce \reasonxl, the first large-scale parallel corpus of cross-domain reasoning traces in five European languages (English, German, French, Italian, and Spanish): over 2.5M aligned samples (11B tokens) per language, each comprising prompt, reasoning trace, and final output, enabling direct supervision of language-specific reasoning.
\ \textcolor{reasoncolor}{\textbf{(b)}} Using \reasonxl, we demonstrate that LLMs can be adapted to reason entirely in a target language through a simple two-stage pipeline of supervised fine-tuning (SFT) and reinforcement learning with verifiable rewards (RLVR), matching or exceeding the base models' English-reasoning performance in two of four languages and recovering most of it in the remaining two.
\end{abstract}
%%%%%%%%%%%%%%%%%%%%%%%%%%%%%%%%%%%%%%%%%%%%%%%%%%%%%%%%%%%%%%%%%%%%%%%%
\section{Introduction}
Recent work shows that training large language models (LLMs) on reasoning data in a small number of languages can generalize to others via cross-lingual transfer \citep{rastogi2025magistral, ghosh2025survey, sutawika2026gained}.
Yet despite these apparent multilingual capabilities, models mostly default to English for intermediate reasoning, even when prompted in other languages \citep{wang-etal-2025-language-mixing}, and consistently achieve higher accuracy in English than elsewhere \citep{saji2025reasoning, yong2025crosslingual}.
For users whose native language is not English, opaque English reasoning traces reduce transparency and trust in the model \citep{aggarwal2025language}, yet existing work has primarily documented these disparities without directly optimizing for reasoning in a target language.

Closing this gap requires two pieces that are currently missing: (1) sufficiently large non-English reasoning data \citep{lai2024mcot, sobhani2025mathmist}, and (2) evidence that language-specific post-training can approach English-level performance \citep{qi2025modelsreasonlanguagecontrolling}.
\textbf{We address both} by introducing, to the best of our knowledge, the first large-scale parallel multilingual reasoning corpus, \reasonxl\footnotemark, and using it in a two-stage post-training pipeline to show that an LLM's reasoning language can be durably redirected to a target language while largely preserving, and in some cases improving, task performance.

\section{Related Work}
\label{sec:related}

% requires: \usepackage{tikz}
%           \usetikzlibrary{arrows.meta, positioning, shapes.geometric, shadows}
\definecolor{pipeRaw}{RGB}{223,223,223}
\definecolor{pipeFilter}{RGB}{255,224,178}
\definecolor{pipeTrans}{RGB}{187,222,251}
\definecolor{pipeFinal}{RGB}{200,230,201}

\begin{table*}[t!]
\centering
\small

\hspace*{-1.6em}%
\begin{minipage}[c]{0.10\textwidth}
\centering
\begin{tikzpicture}[
  stage/.style={draw, rounded corners=2pt, align=center, text width=1.55cm,
                minimum height=0.8cm, font=\scriptsize, inner sep=2pt,
                drop shadow={shadow xshift=0.4pt, shadow yshift=-0.6pt, opacity=0.35}},
  funnel/.style={stage, trapezium, trapezium left angle=75, trapezium right angle=105,
                 trapezium stretches=true, minimum width=1.5cm},
  arr/.style={-{Stealth[length=1.8mm]}, thick, shorten >=1pt, shorten <=1pt}
]
\node[stage,  fill=pipeRaw]                          (raw)   {\faDatabase\\[1pt]English\\Corpus};

\node[funnel, fill=pipeFilter, below=9pt of raw]      (pre)   {\faFilter\ Pre-\\Filtering};

\node[stage,  fill=pipeTrans,  below=9pt of pre]      (trans) {\faLanguage\\[1pt]Translate\\\texttt{\tiny Qwen3-32B}};

\node[funnel, fill=pipeFinal,  below=9pt of trans]    (post)  {\faFilter\ Post-\\Filtering};

\draw[arr] (raw)   -- (pre);
\draw[arr] (pre)   -- (trans);
\draw[arr] (trans) -- (post);
\end{tikzpicture}
\end{minipage}%
\hspace{0.05\textwidth}
\begin{minipage}[c]{0.46\textwidth}
\vspace{0pt}
\raggedright

\begin{minipage}[t]{0.90\linewidth}
\vspace{0pt}
\centering
\setlength{\tabcolsep}{4pt}
\begin{tabular}{
l
S[table-format=2.2,table-number-alignment=center]
S[table-format=4.1,table-number-alignment=center]
S[table-format=3.1,table-number-alignment=center]
S[table-format=4.1,table-number-alignment=center]
}
\toprule
& {\textbf{\faDatabase\ Tokens}} & \multicolumn{3}{c}{\faRuler\ \textit{Seq.\ Length (avg.)}} \\
\cmidrule(lr){3-5}
\textbf{Lang} & {\textbf{Total (B)}} & {\textbf{Total}} & {\textbf{Input}} & {\textbf{Output}} \\
\midrule
\rowcolor{rowgray}
\textsc{en} & 11.17 & 4398.9 & 399.9 & 3998.9 \\
\textsc{de} & 10.66 & 4200.9 & 468.0 & 3733.0 \\
\rowcolor{rowgray}
\textsc{fr} & 10.77 & 4243.0 & 459.8 & 3783.2 \\
\textsc{es} & 10.54 & 4152.3 & 447.8 & 3704.6 \\
\rowcolor{rowgray}
\textsc{it} & 10.70 & 4217.1 & 461.6 & 3755.5 \\
\midrule
\textbf{Total} & {\bfseries 53.85} & \multicolumn{3}{c}{---} \\
\bottomrule
\end{tabular}
\end{minipage}

\vspace{1.4em}

\begin{minipage}[t]{0.90\linewidth}
\vspace{0pt}
\centering
\setlength{\tabcolsep}{4pt}
\begin{tabular}{p{0.50\linewidth}l r}
\toprule
\textbf{Dataset} & \textbf{Config} & \textbf{Samples} \\
\midrule

\rowcolor{rowgray}
Cascade-SFT-Stage-2$^{1}$ & \tmid general & 611,367 \\
\rowcolor{rowgray}
                          & \tend math    & 103,222 \\

Dolci-Think-SFT-7B$^{2}$  & \tend --- & 510,466 \\

\rowcolor{rowgray}
Cascade-SFT-Stage-1$^{1}$ & \tmid general & 314,099 \\
\rowcolor{rowgray}
                          & \tmid code    & 130,614 \\
\rowcolor{rowgray}
                          & \tmid math    & 176,276 \\
\rowcolor{rowgray}
                          & \tend science &  68,614 \\

Llama-Nemotron-PTD$^{3}$  & \tend science & 484,703 \\

\rowcolor{rowgray}
Nemotron-Science-v1$^{4}$ & \tend --- & 100,419 \\

Nemotron-IF-Chat-v1$^{4}$ & \tend --- &  38,670 \\

\midrule
\textbf{Total} & & \textbf{2,538,450} \\
\bottomrule
\end{tabular}
\end{minipage}

\end{minipage}\hspace{0.002\textwidth}
\begin{minipage}[c]{0.33\linewidth}
\vspace{0pt}
\raggedright
\setlength{\tabcolsep}{4pt}
\begin{tabular}{
>{\raggedright\arraybackslash}p{0.42\linewidth}
>{\raggedright\arraybackslash}p{0.36\linewidth}
S[table-format=2.1,table-number-alignment=center]
}
\toprule
\textbf{\faSlidersH\ Property} & \textbf{Value} & {\textbf{\%}} \\
\midrule

Tech.\ Content & \tmid Math      & 24.9 \\
                  & \tmid Non~tech.   & 24.5 \\
                  & \tmid Scientific      & 23.5 \\
                  & \tmid Basic~tech. & 16.4 \\
                  & \tmid Code      &  8.1 \\
                  & \tmid Engineering     &  1.4 \\
                  & \tend Data      &  1.3 \\
\midrule

\rowcolor{rowgray}
Info.\ Density & \tmid dense     & 60.2 \\
\rowcolor{rowgray}
               & \tmid adequate  & 37.2 \\
\rowcolor{rowgray}
               & \tend moderate  &  2.6 \\
% \rowcolor{rowgray}
%                & \tend thin      &  0.1 \\
\midrule

Edu.\ Value    & \tmid high      & 55.3 \\
               & \tmid moderate  & 33.8 \\
               & \tmid basic     &  7.9 \\
               & \tend minimal   &  3.0 \\
               % & \tend none      &  1.2 \\
\midrule

\rowcolor{rowgray}
Audience       & \tmid advanced  & 42.8 \\
\rowcolor{rowgray}
               & \tmid general   & 33.0 \\
\rowcolor{rowgray}
               & \tmid beginner  & 17.8 \\
\rowcolor{rowgray}
               & \tend expert    &  6.5 \\
% \rowcolor{rowgray}
%                & \tmid children  &  1.3 \\
% \rowcolor{rowgray}
%                & \tend youth     &  2.2 \\
% \midrule

% Content Length & \tmid brief        & 49.4 \\
%                & \tmid moderate     & 43.1 \\
%                & \tend substantial  &  7.5 \\

\bottomrule
\end{tabular}

\vspace{1.4em}

\centering
\setlength{\tabcolsep}{4pt}
\begin{tabular}{lc}
\toprule
\textbf{Model/Task} & \textit{Avg} \\
\midrule
\rowcolor{rowgray}SmolLM3-3 \hflogo        & $74.3$ \\
\quad + \texttt{EN SFT}                    & $\boldsymbol{77.0}$ \\
\bottomrule
\end{tabular}

\end{minipage}

\caption{Dataset construction pipeline (far left), per-language corpus size and sequence lengths (top-middle), composition (bottom-middle), and property distributions (top-right). 
% Input length is computed as total tokens minus target tokens.
Dataset sources: $^{1}$\cite{Nemotron_Cascade_Scaling_Cascaded_Reinforcement_Learning}, $^{2}$\cite{olmo2025olmo3}, $^{3}$\cite{bercovich2025llamanemotronefficientreasoningmodels}, $^{4}$\cite{nvidia_nemotron_3_2025}. Bottom-right: English performance of \texttt{SmolLM3-3B} pre- and post-SFT: \textbf{consistent gains demonstrate the dataset's potential to push strong baselines further.}}
\label{tab:multilingual_dataset_stats}
\end{table*}

\footnotetext{\textbf{XL} stands for \textbf{cross-l}ingual, and reflects the scale of the corpus. The data is made available on \href{https://huggingface.co/datasets/toroe/ReasonXL-SFT}{HuggingFace}.
% : \href{https://huggingface.co/collections/DGurgurov/reasonxl-models-and-data}{DGurgurov/reasonxl}
}

\textbf{Multilingual Reasoning.}
Despite rapid progress in multilingual LLMs \citep{dang2024ayaexpansecombiningresearch, qwen3technicalreport}, models still default to English for intermediate reasoning and achieve higher accuracy in English than in other languages \citep{wang-etal-2025-language-mixing, tam2025language, saji2025reasoning}.
Approaches to steering the reasoning language fall into three categories.
\textit{Translation-based methods} use English as a pivot \citep{shi2022languagemodelsmultilingualchainofthought, ahuja2023mega, etxaniz2024multilingual, zhang2024plug, ko2025understand, yoon2024langbridge, she2024mapo, fan2025slam}.
\textit{Prompt-based methods} control the reasoning language without fine-tuning, but trade accuracy for compliance \citep{huang2023not, yong2025crosslingual, qi2025modelsreasonlanguagecontrolling}.
\textit{Post-training approaches} align the reasoning language more effectively, even with limited data, though a performance gap to English remains that test-time scaling narrows only inconsistently \citep{qi2025modelsreasonlanguagecontrolling, son2025linguisticgeneralizabilitytesttimescaling}.
Progress across all three is constrained by the scarcity of large-scale non-English reasoning data: existing corpora are English-prevalent, limited in scale, or lack intermediate reasoning traces \citep{lai2024mcot, sobhani2025mathmist, olmo2025olmo3, guha2026openthoughts, ghosh2025survey}.

\textbf{The Case for Non-English Reasoning.}
In-language reasoning lets users inspect reasoning traces directly, improving trust and preserving linguistic and cultural nuance \citep{aggarwal2025language}; routing through English instead introduces ``lost in translation'' errors that distort problem semantics \citep{saji2025reasoning} and limits oversight in multilingual deployments \citep{qi2025modelsreasonlanguagecontrolling}.
Unlike prior post-training work operating on limited data, we pair a purpose-built corpus with the dedicated adaptation: \reasonxl{} closes the data gap with 2.5M cross-domain traces per language (Section~\ref{sec:meth}), and post-training on it shifts the reasoning language while largely preserving performance (Section~\ref{sec:ver}).
\begin{table*}[h!]
\centering
\small
\setlength{\tabcolsep}{5pt}
\begin{tabular}{c|ll|ccc:c|ccc:c}
\toprule
& & & \multicolumn{4}{c|}{\hflogo\ \textsc{SmolLM3-3B}} & \multicolumn{4}{c}{\openthinker\ \textsc{OpenThinker3-7B}} \\
\cmidrule(lr){4-7} \cmidrule(lr){8-11}
& \textbf{Lang} & \textbf{Method} & \faFlask\ \textit{Sci.} & \faLanguage\ \textit{Lang.} & \textit{Ov.} & \textit{\%TL} & \faFlask\ \textit{Sci.} & \faLanguage\ \textit{Lang.} & \textit{Ov.} & \textit{\%TL} \\
\midrule
& \textsc{de} & Base & \cellcolor{rowgray}$61.6$ & \cellcolor{rowgray}$65.5$ & \cellcolor{rowgray}$63.5$ & \cellcolor{rowgray}$0.0$ & \cellcolor{rowgray}$\boldsymbol{61.3}$ & \cellcolor{rowgray}$58.4$ & \cellcolor{rowgray}$59.9$ & \cellcolor{rowgray}$0.0$ \\
&  & \quad + SFT & $50.9$ & $60.7$ & $55.8$ & $96.5$ & $33.7$ & $51.6$ & $42.6$ & $87.3$ \\
&  & \quad + GRPO & \cellcolor{lightolivegreen-3!50}$\boldsymbol{61.6}$ & \cellcolor{lightolivegreen-3!50}$\boldsymbol{65.9}$ & \cellcolor{lightolivegreen-3!50}$\boldsymbol{63.7}$ & \cellcolor{lightolivegreen-3!50}$100.0$ & \cellcolor{lightolivegreen-3!50}$57.7$ & \cellcolor{lightolivegreen-3!50}$\boldsymbol{63.8}$ & \cellcolor{lightolivegreen-3!50}$\boldsymbol{60.7}$ & \cellcolor{lightolivegreen-3!50}$100.0$ \\
\cmidrule{2-11}
& \textsc{it} & Base & \cellcolor{rowgray}$62.2$ & \cellcolor{rowgray}$67.6$ & \cellcolor{rowgray}$64.9$ & \cellcolor{rowgray}$0.0$ & \cellcolor{rowgray}$\boldsymbol{61.3}$ & \cellcolor{rowgray}$65.4$ & \cellcolor{rowgray}$63.3$ & \cellcolor{rowgray}$0.0$ \\
&  & \quad + SFT & $58.1$ & $71.2$ & $64.6$ & $96.2$ & $35.8$ & $52.6$ & $44.2$ & $92.0$ \\
&  & \quad + GRPO & \cellcolor{lightolivegreen-3!50}$\boldsymbol{65.1}$ & \cellcolor{lightolivegreen-3!50}$\boldsymbol{75.2}$ & \cellcolor{lightolivegreen-3!50}$\boldsymbol{70.2}$ & \cellcolor{lightolivegreen-3!50}$100.0$ & \cellcolor{lightolivegreen-3!50}$58.2$ & \cellcolor{lightolivegreen-3!50}$\boldsymbol{71.3}$ & \cellcolor{lightolivegreen-3!50}$\boldsymbol{64.7}$ & \cellcolor{lightolivegreen-3!50}$100.0$ \\
\cmidrule{2-11}
& \textsc{es} & Base & \cellcolor{rowgray}$\boldsymbol{60.8}$ & \cellcolor{rowgray}$71.3$ & \cellcolor{rowgray}$\boldsymbol{66.0}$ & \cellcolor{rowgray}$0.0$ & \cellcolor{rowgray}$\boldsymbol{69.2}$ & \cellcolor{rowgray}$\boldsymbol{70.7}$ & \cellcolor{rowgray}$\boldsymbol{69.9}$ & \cellcolor{rowgray}$0.0$ \\
&  & \quad + SFT & $56.0$ & $69.1$ & $62.5$ & $78.3$ & $42.0$ & $61.3$ & $51.7$ & $72.5$ \\
&  & \quad + GRPO & \cellcolor{lightolivegreen-3!50}$54.1$ & \cellcolor{lightolivegreen-3!50}$\boldsymbol{72.0}$ & \cellcolor{lightolivegreen-3!50}$63.0$ & \cellcolor{lightolivegreen-3!50}$100.0$ & \cellcolor{lightolivegreen-3!50}$48.4$ & \cellcolor{lightolivegreen-3!50}$66.7$ & \cellcolor{lightolivegreen-3!50}$57.6$ & \cellcolor{lightolivegreen-3!50}$100.0$ \\
\cmidrule{2-11}
& \textsc{fr} & Base & \cellcolor{rowgray}$\boldsymbol{59.7}$ & \cellcolor{rowgray}$\boldsymbol{69.3}$ & \cellcolor{rowgray}$\boldsymbol{64.5}$ & \cellcolor{rowgray}$0.0$ & \cellcolor{rowgray}$\boldsymbol{67.3}$ & \cellcolor{rowgray}$\boldsymbol{68.5}$ & \cellcolor{rowgray}$\boldsymbol{67.9}$ & \cellcolor{rowgray}$0.0$ \\
&  & \quad + SFT & $55.2$ & $66.6$ & $60.9$ & $95.8$ & $33.6$ & $57.1$ & $45.4$ & $80.9$ \\
&  & \quad + GRPO & \cellcolor{lightolivegreen-3!50}$55.5$ & \cellcolor{lightolivegreen-3!50}$68.1$ & \cellcolor{lightolivegreen-3!50}$61.8$ & \cellcolor{lightolivegreen-3!50}$100.0$ & \cellcolor{lightolivegreen-3!50}$52.0$ & \cellcolor{lightolivegreen-3!50}$65.4$ & \cellcolor{lightolivegreen-3!50}$58.7$ & \cellcolor{lightolivegreen-3!50}$100.0$ \\
\bottomrule
\end{tabular}
\caption{Domain-averaged performance on target-language tasks. \textbf{SFT steers the model to reason in the target language; GRPO recovers performance while preserving compliance.} \textit{Sci.}: MATH100/GPQA/MGSM; \textit{Lang.}: Belebele/GMMLU/INCLUDE; \textit{Ov.}: all six tasks, each averaged over 5 runs.}
\label{tab:multilingual-results-domain}
\end{table*}

\section{Dataset Construction}
\label{sec:meth}

\label{sec:data-creation}
We construct \reasonxl{} by annotating, curating, and translating a multilingual cross-domain reasoning corpus.
The English source data is sampled from six existing open resources \citep{olmo2025olmo3, nvidia_nemotron_3_2025, bercovich2025llamanemotronefficientreasoningmodels, Nemotron_Cascade_Scaling_Cascaded_Reinforcement_Learning} and translated into German, French, Spanish, and Italian, yielding a parallel corpus in five languages (composition in Table~\ref{tab:multilingual_dataset_stats}, bottom-middle).

\textbf{Data Annotation and Sampling.}
We annotate all English samples with \texttt{Propella-1-4B} \citep{idahl2026propella1multipropertydocumentannotation}, a small multilingual LLM for scalable document annotation that scores 18 properties across six categories, enabling a multi-stage filtering and balancing pipeline for safety, quality, and domain diversity.
First, mandatory \textit{integrity and safety constraints} retain only documents labeled safe, free of personally identifiable information, and complete, while removing non-informational document types and requiring the presence of reasoning indicators.
Next, \textit{domain-dependent quality thresholds} hold technically demanding content (e.g., math or code) to stricter criteria, such as evergreen, information-dense, educationally valuable, and of excellent overall quality, while other domains use a relaxed threshold to maintain topical coverage.
Finally, \textit{class-aware downsampling} mitigates overrepresentation of certain technical categories, yielding a balanced distribution across content classes (details in Appendix~\ref{app:dataset_filtering}).

\textbf{Translation Protocol.}\label{para:translation} 
Each sample comprises three independently translated components: user input, reasoning trace, and final output.
We translate with \texttt{Qwen3-32B} \citep{qwen3technicalreport}, selected for its strong multilingual performance and because, in preliminary experiments, LLM-based translation preserved structural elements such as equations and reasoning traces better than standard MT systems.
A detailed system prompt instructs the model to preserve technical terminology and mathematical notation while adapting tone and cultural references to the target language (prompt in Appendix~\ref{app:translation-prompt}, sampling parameters in Appendix~\ref{app:translation-hp}, sample translations in Appendix~\ref{app:dataset_samples}).

\textbf{Post-Translation Filtering.}
After translation, we filter each language split for degenerate or unfaithful translations.
Language identity of prompt, reasoning trace, and output is verified with fastText \citep{joulin2016bag} over prose only (excluding code, math, and markup), exempting samples whose prompt explicitly requests another language (e.g., translation exercises).
We further remove degenerate text (repetition loops, spuriously upper-cased passages, encoding corruption), broken structure (unbalanced \texttt{<think>} tags, malformed \verb|\boxed{}| expressions, truncated conversations), and final answers that no longer match the English source.
Samples where the translator leaked its own instructions into the output are repaired by excising the leaked span where possible, and discarded otherwise.
As a control, the same checks run on the untranslated English source, where translation-specific issues occur at near-zero rates.
Since each language retains a slightly different subset, we finally intersect surviving sample IDs across all five languages and de-duplicate on the English side, keeping the corpus parallel (full criteria and per-language statistics in Appendix~\ref{app:dataset_filtering_post}).

\textbf{Quality Verification.} 
The final English source dataset (Table~\ref{tab:multilingual_dataset_stats}, top-right) comprises a balanced mix of technical domains and high information-density samples.
Since the Propella-1 annotations capture semantic properties, they should transfer to all parallel translations without re-annotation; we nevertheless test this assumption by re-annotating the full target-language splits (DE, FR, IT, ES) with Propella-1, recovering the source property profile in every language (Appendix~\ref{app:data-quality}).
Because the labels capture semantic and structural content rather than surface form, this preservation indicates that translation introduced no semantic collapse, dropped reasoning steps, or degenerate text.

\section{Corpus Verification}
\label{sec:ver}
\subsection{Experimental Setup}

We adopt a two-stage training pipeline applied independently to each target language, using \texttt{SmolLM3-3B} \citep{bakouch2025smollm3} and \texttt{OpenThinker3-7B} \citep{guha2026openthoughts} as base models, both supporting all languages in our corpus.

\textbf{Stage 1: Supervised Fine-Tuning (SFT)}
shifts the model's reasoning language from English to the target language by exposing it to the distribution of in-language reasoning traces.
For each language, we tune each base model on the corresponding split of our parallel corpus with completion-only loss (user and system tokens masked), packing chat-formatted sequences to 16{,}384 tokens for 2 epochs (hyperparameters in Appendix~\ref{app:sft-hyperparams}).

\textbf{Stage 2: RL with Dr.\ GRPO}
recovers the reasoning quality lost during SFT while retaining the target-language compliance it establishes.
Starting from each language-specific SFT checkpoint, we apply Dr.\ GRPO \citep{liu2025understandingr1zeroliketrainingcritical} on a per-language set of 20K verifiable prompts, drawn from translated subsets of MATH \citep{hendrycksmath2021} and GSM8K \citep{cobbe2021gsm8k} via the pipeline of Section~\ref{sec:meth}.
Each prompt yields $G{=}8$ candidate completions, scored by a composite reward combining \textbf{accuracy} (symbolic equivalence of the predicted \texttt{\textbackslash boxed\{\}} answer with the ground truth), \textbf{language identity} (fastText scores over reasoning trace and output), \textbf{format compliance} (\texttt{<think>} and \texttt{\textbackslash boxed\{\}} usage), and a \textbf{repetition penalty} preventing length-hacking; weights and full definitions are in Appendix~\ref{app:grpo-details}.

\textbf{Evaluation.}
In-domain evaluation uses three target-language science benchmarks, given their topical overlap with the SFT and RL data: \textit{MGSM} \citep{shi2022languagemodelsmultilingualchainofthought}, \textit{MT-Math100} \citep{son2025linguisticgeneralizabilitytesttimescaling}, and \textit{GPQA Diamond} \citep{rein2024gpqa, qi2025modelsreasonlanguagecontrolling}.
Three further benchmarks measure out-of-domain generalization: \textit{Global MMLU Lite} \citep{singh2024globalmmluunderstandingaddressing} for world knowledge, \textit{Belebele} \citep{Bandarkar_2024} for reading comprehension, and \textit{INCLUDE} \citep{romanou2024includeevaluatingmultilinguallanguage} for regional academic and professional knowledge.
All evaluations use vLLM \citep{kwon2023efficient} at temperature $0.6$, top-$p$ $0.95$, averaged over 5 runs, with answers requested inside \verb|\boxed{}| in the target language; adapted models use the RL-stage system prompt, base and SFT models none (s. Appendix~\ref{app:eval-details}).

\subsection{Adaptation Results}
\label{sec:adapt}

We first validate the English corpus itself: SFT on the English split alone yields average gains across all benchmarks (Table~\ref{tab:multilingual_dataset_stats}, bottom-right), improving strong baselines even in English.

Table~\ref{tab:multilingual-results-domain} presents the core finding of this work.
Across all four target languages, both base models reason exclusively in English (\% of target language, TL\,=\,0, measured with FastText \cite{joulin2016bag}).
SFT mostly redirects reasoning into the target language (\%TL\,>\,70), but at a steep cost: science performance drops by 4-11 points for \texttt{SmolLM3-3B} and 25-34 points for \texttt{OpenThinker3-7B}, consistent with prior observations that supervised language forcing alone degrades reasoning quality~\citep{qi2025modelsreasonlanguagecontrolling}.
GRPO then recovers performance while fully establishing target-language compliance: in German and Italian, the final models match or outperform the base model overall for both backbones, and in Spanish and French they recover most of the SFT degradation.
Out-of-domain benchmarks likewise improve after GRPO in German and Italian, indicating that the RL stage strengthens general in-language capabilities rather than overfitting to the science domain.

To our knowledge, this is the \textbf{first demonstration that the reasoning language of an LLM can be shifted entirely to another language while matching its English-reasoning performance}.
Beyond feasibility, these results serve as a behavioral check on data quality, complementing the annotation-level check of Section~\ref{sec:meth}: SFT reliably induces in-language reasoning and GRPO recovers accuracy up to or beyond the base model in half the settings, neither expected if translation had systematically corrupted the traces.

\section{Discussion and Conclusion}

\field{inputcolor}{\textbf{(a)}} \textbf{An Open Resource for Multilingual Reasoning.}
\reasonxl\ addresses the data scarcity identified in Section~\ref{sec:related} with over 2.5M aligned samples across five languages, each enriched with \texttt{Propella-1} metadata; we release it publicly to support the training of multilingual LLMs, cross-lingual transfer studies, analysis of reasoning traces, curriculum design for multilingual models, and evaluation of language-specific chain-of-thought faithfulness.

\field{reasoncolor}{\textbf{(b)}} \textbf{Language-Specific Reasoning Adaptation.}
Our experiments are an initial feasibility demonstration, not a fully optimized pipeline: two base models, a fixed two-stage recipe, and no exhaustive tuning of hyperparameters, reward weights, or data filtering.
Even this simple recipe matches or exceeds English-reasoning performance in German and Italian, suggesting clear headroom; larger models, advanced RL strategies, and mixed-language curricula may provide even stronger results across languages.

In sum, a sufficiently large, well-structured corpus allows an LLM's reasoning language to be shifted entirely to a target language via straightforward post-training while largely preserving performance, challenging the assumption that English is necessary for high-quality reasoning.
\graphicspath{{assets/}}
%%%%%%%%%%%%%%%%%%%%%%%%%%%%%%%%%%%%%%%%%%%%%%%%%%%%%%%%%%%%%%%%%%%%%%%%

\section*{Limitations}

Our experiments are restricted to two models of 3B and 7B parameters and five typologically related European languages; generalization to larger scales and more languages remains to be verified.
The translations, while produced by a strong multilingual model, were not validated by human annotators, and subtle translationese artifacts could influence training dynamics.
Finally, adaptation does not fully close the gap to the base models in Spanish and French; we use a single shared training configuration across languages, and per-language tuning of hyperparameters and reward weights may be required, which we leave to future work.
% Finally, our adaptation experiments target one language at a time; whether multilingual joint training provides different trade-offs between language compliance, reasoning quality, and cross-lingual transfer remains an open question.

% \section*{Acknowledgments}
% This work is funded by the German Federal Ministry for Economic Affairs and Energy (BMWE) as part of the project “Souveräne KI für Europa (SOOFI)” (13IPC040H), the German Federal Ministry of Research, Technology and Space (BMFTR) as part of the project TRAILS (01IW24005), the German Federal Ministry of Education and Research (BMBF)  under the project AI4SCM (01|S23015A), the Deutsche Forschungsgemeinschaft  (DFG, German Research Foundation) Project-ID 528483508 - FIP 12 and the European Union under the grant project COMFORT - Improving Urologic Cancer Care with Artificial Intelligence Solutions (101079894).

% \section*{Ethics Statement}
% Authors can add an optional ethics statement to the paper. 
% For papers that touch on ethical issues, this section will be evaluated as part of the review process. The ethics statement should come at the end of the paper. It does not count toward the page limit, but should not be more than 1 page. 

\bibliography{custom}

\newpage
\appendix
\section*{Appendix}

\section{Translation Prompt}
\label{app:translation-prompt}
\begin{figure}[h]
\begin{tcolorbox}[colback=gray!5, colframe=gray!50, boxrule=0.5pt]
\small
\begin{verbatim}
LANGUAGE_CONFIG = {
"german": {"code": "de", "name": "German", 
        "formal_context": "Use formal German 
    (Sie) for professional/technical content"},
"spanish": {"code": "es", "name": "Spanish", 
        "formal_context": "Use neutral Spanish 
        suitable for international audiences"},
"french": {"code": "fr", "name": "French", 
    "formal_context": "Use standard French 
    with appropriate formality"},
"italian": {"code": "it", "name": "Italian", 
    "formal_context": "Use standard Italian 
    with professional tone"},
}

SYSTEM: You are a professional translator 
specializing in technical and 
educational content. Translate the following 
{field} text into {language}.

CRITICAL INSTRUCTIONS:
1. Output ONLY the translated text
2. Preserve ALL technical terms, code snippets, 
mathematical notation, 
and formatting exactly
3. Maintain the same tone, style, and formality
4. {language-specific formality guidance}
5. For code: Keep variable/function 
    names in English
6. For math: Preserve LaTeX notation unchanged
7. Adapt examples and cultural 
    references appropriately
8. Maintain terminology consistency throughout

USER: TEXT TO TRANSLATE:
{text}

TRANSLATION:
\end{verbatim}
\end{tcolorbox}
\caption{Translation prompt template. \texttt{\{field\}} is one of: input, thinking/reasoning, or output.}
\label{fig:translation-prompt}
\end{figure}

\clearpage

\begin{table*}[h!]
\section{Dataset Samples}
\label{app:dataset_samples}
\centering
\small
\begin{tabularx}{\textwidth}{@{} c X @{}}
\toprule
\textbf{Lang} & \textbf{Sample} \\
\midrule
\rowcolor{rowgray}
\textsc{en} 
  & \field{inputcolor}{Input:} Which of the following statements is true about the singularity at the center of a black hole? \newline
    \field{reasoncolor}{Reasoning:} Okay, I need to figure out which of these options is correct about the singularity at the center of a black hole. \newline
    \field{outputcolor}{Output:} The answer is (C). **Explanation:**  A singularity at the center of a black hole is a region where the spacetime curvature becomes infinite.. \\[0.4em]
    \addlinespace[0.1em]
\textsc{de} 
  & \field{inputcolor}{Input:} Welcher der folgenden Aussagen ist in Bezug auf die Singularität im Zentrum eines Schwarzen Lochs wahr? \newline
    \field{reasoncolor}{Reasoning:} Okay, ich muss herausfinden, welche dieser Optionen in Bezug auf die Singularität im Zentrum eines Schwarzen Lochs korrekt ist. \newline
    \field{outputcolor}{Output:} Die Antwort ist (C). **Erklärung:**  Eine Singularität im Zentrum eines Schwarzen Lochs ist eine Region.. \\[0.4em]
    \addlinespace[0.1em]
\textsc{es} 
  & \field{inputcolor}{Input:} ¿Cuál de las siguientes afirmaciones es verdadera sobre la singularidad en el centro de un agujero negro? \newline
    \field{reasoncolor}{Reasoning:} Bueno, necesito averiguar cuál de estas opciones es correcta acerca de la singularidad en el centro de un agujero negro. \newline
    \field{outputcolor}{Output:} La respuesta es (C). **Explicación:**  Una singularidad en el centro de un agujero negro es una región donde la curvatura del espacio-tiempo se vuelve infinita.. \\[0.4em]
    \addlinespace[0.1em]
\textsc{fr} 
  & \field{inputcolor}{Input:} Laquelle des affirmations suivantes est vraie concernant la singularité au centre d'un trou noir ? \newline
    \field{reasoncolor}{Reasoning:} D'accord, je dois déterminer laquelle de ces options est correcte concernant la singularité au centre d'un trou noir. \newline
    \field{outputcolor}{Output:} La réponse est (C). **Explication :**  Une singularité au centre d'un trou noir est une région où la courbure de l'espace-temps devient infinie.. \\[0.2em]
    \addlinespace[0.1em]
\textsc{it} 
  & \field{inputcolor}{Input:} Quali delle seguenti affermazioni è vera riguardo alla singolarità al centro di un buco nero? \newline
    \field{reasoncolor}{Reasoning:} Okay, devo capire quale di queste opzioni è corretta riguardo alla singolarità al centro di un buco nero. \newline
    \field{outputcolor}{Output:} La risposta è (C).**Spiegazione:**  Una singolarità al centro di un buco nero è una regione in cui la curvatura dello spaziotempo diventa infinita... \\[0.2em]
    \addlinespace[0.1em]
\bottomrule
\end{tabularx}
\caption{Example translations from the parallel \reasonxl\ corpus.}
\label{tab:translation-examples}
\end{table*}
\clearpage

\section{Pre-Translation Filtering and Sampling}
\label{app:dataset_filtering}

\subsection{Mandatory Safety and Integrity Filters}
All documents must be labeled \texttt{safe}, \texttt{no\_pii}, \texttt{complete}, and \texttt{complete\_content}, must show reasoning indicators (not \texttt{none}), and must be free of commercial bias.
We further remove non-informational or transactional document types (\texttt{press\_release}, \texttt{boilerplate}, \texttt{news\_report}, \texttt{transactional}, \texttt{legal\_document}), exclude low-frequency business sectors (\texttt{other}, \texttt{mining\_resources}, \texttt{wholesale\_distribution}), and retain only documents with content length \texttt{brief}, \texttt{moderate}, or \texttt{substantial}.

\subsection{Domain-Dependent Quality Filtering}
Documents in the strict technical domains \texttt{math\_heavy} and \texttt{code\_heavy} must additionally be \texttt{evergreen}, \texttt{dense}, of \texttt{high} or \texttt{moderate} educational value, and of \texttt{excellent} content quality.
All other domains use a relaxed constraint: content quality must be \texttt{excellent}, \texttt{good}, or \texttt{adequate}.

\subsection{Class-Balanced Sampling}

After filtering, we perform class-aware downsampling to avoid overrepresentation of certain technical categories.

Table~\ref{tab:sampling_ratios} shows the sampling ratios applied to each technical content class.

\begin{table}[h]
\small
\centering
\begin{tabular}{lc}
\hline
Technical Content Class & Sampling Ratio \\
\hline
\texttt{code\_heavy\_math\_heavy} & 0.60 \\
\texttt{math\_heavy} & 0.30 \\
\texttt{non\_technical} & 0.50 \\
\texttt{basic\_technical} & 0.80 \\
\hline
\end{tabular}
\caption{Class-aware sampling ratios used during dataset balancing.}
\label{tab:sampling_ratios}
\end{table}

For each class $c$, we randomly sample a subset of documents proportional to the class-specific ratio. Documents belonging to classes not listed in Table~\ref{tab:sampling_ratios} are retained without downsampling.

\subsection{Final Dataset Construction}
\label{app:final_ds}
The final dataset combines the downsampled documents with all documents from non-reduced classes.
The resulting distribution (Table~\ref{tab:multilingual_dataset_stats}, top-right) is balanced across technical content classes and strongly skewed toward information-rich, educationally valuable material, reflecting the curation pipeline's priorities.

\section{Translation Hyperparameters}
\label{app:translation-hp}
Table ~\ref{tab:translation_sampling_params} shows the Qwen3-32B sampling parameters.
\begin{table}[h]
\centering
\small
\setlength{\tabcolsep}{6pt}
\begin{tabular}{l c}
\toprule
\textbf{Category} & \textbf{Configuration} \\
\midrule
Decoding Strategy & Nucleus Sampling \\
Temperature       & 0.1 \\
Top-$p$           & 1.0 \\
Top-$k$           & $-1$ \\
Max Length        & 6{,}144 tokens \\
\bottomrule
\end{tabular}
\caption{Decoding and generation parameters used for translation.}
\label{tab:translation_sampling_params}
\end{table}

\section{Post-Translation Filtering}
\label{app:dataset_filtering_post}

All checks are deterministic and computed in a single pass over each split.
Samples are matched across languages by an ID composed of the sample UID, source, and dataset name; the English sample with the same ID serves as the reference for each translation.
 
\subsection{Language Checks}
\label{app:post-lid}
Before language identification, we strip code, math, LaTeX commands, and URLs, since these are deliberately left untranslated (Appendix~\ref{app:translation-prompt}), and run the fastText \texttt{lid.176} model \citep{joulin2016bag} on the remaining prose.
Since the most common translation failure is a correctly translated prompt with an untranslated reasoning trace, we check the language separately for the user prompt, the reasoning trace, the final answer, and the whole assistant output (confidence ${\geq}0.60$), as well as for individual paragraphs (confidence ${\geq}0.70$); a sample counts as partially untranslated when ${\geq}25\%$ of its paragraphs are confidently identified as a different language.
A mismatch alone is not sufficient reason for removal, since the corpus legitimately contains translation exercises and prompts explicitly asking for another language, and filtering on mismatch alone would delete this task category.
We therefore remove a sample only when its prompt is in the target language \emph{and} contains no explicit language request (matched against a list of translate/rewrite verbs, language names, and demonyms).
A separate check catches a similar failure, where the answer was translated but the prompt was left in English.
 
\subsection{Degenerate Text}
\label{app:post-degeneration}
We remove repetition loops (a run of ${\geq}30$ identical tokens, ${\geq}15$ identical $n$-grams for $n\in[2,5]$, or a zlib compression ratio ${\geq}18$ on ${\geq}200$ tokens), samples containing mojibake or Unicode replacement characters, and samples whose reasoning trace or final answer is ${\geq}60\%$ upper-cased.
Upper-cased passages are a translation artifact: under the identical check, they occur in 0.22\% (FR) to 5.42\% (DE) of translated samples but only 0.005\% of the English source.
We remove these outright, since reliable recasing (particularly for German, where nouns are capitalized) would amount to re-generating the text.

\subsection{Leaked Translator Instructions}
\label{app:post-leak}
In a few percent of samples, the translator copied its own instruction list (Appendix~\ref{app:translation-prompt}) into the output, usually at the start of the reasoning trace.
Searching for literal marker strings does not work here, as the block appears \emph{in the target language} and is rendered inconsistently from sample to sample, including upper-cased and garbled variants.
We instead recognize the leak by its shape: an ascending numbered list of ${\geq}6$ items mentioning both LaTeX and a translate-verb stem, which is characteristic of our translation prompt and essentially never occurs in genuine reasoning.
On a 200K-sample German subset, this recognizes every sample found by a literal marker search plus 30.4\% additional (mostly upper-cased) variants.
Since the block is a contiguous, clearly delimited span, we remove it and keep the sample, provided the remaining reasoning trace is non-trivial (${\geq}40$ characters), the final answer is non-empty, and the \texttt{<think>} tags stay balanced; otherwise the leak replaced the reasoning trace entirely and the sample is discarded (in the full German split, 36{,}452 samples repaired and 18{,}487 discarded).

\subsection{Structure and Answer Consistency}
\label{app:post-structural}
We require valid conversations (starting with a user turn, ending with an assistant turn, no repeated roles or empty turns), balanced \texttt{<think>}/\texttt{</think>} tags followed by a non-empty final answer, closed code fences, well-formed \verb|\boxed{}| expressions, and no truncation markers.
Differences that would propagate to all languages through the parallel intersection are normalized away first: content-free leading \texttt{system} turns exist only in the source split, and treating them as a defect would have deleted the affected samples from all five languages.
In addition, we extract the final answer from each translation (last \verb|\boxed{}| expression, localized \textit{answer:} patterns, option letters, or short numeric literals), normalize it, and remove samples whose answer mismatches the one extracted identically from the English reference.
This applies only when both sides yield a concrete answer, but catches translation of the answer token itself (e.g., \texttt{\textbackslash text\{nein\}} vs.\ \texttt{\textbackslash text\{no\}}), which no monolingual check can detect.
Removed samples are stored separately together with the reason for their removal, so every exclusion remains traceable.

\subsection{The English Source as a Control}
\label{app:post-control}
Since the corpus contains its own untranslated source, we run every check on the English split under identical settings, giving a baseline rate for each check (Table~\ref{tab:post-control}).
The instruction-leak check returns exactly zero on English, where the leak cannot occur by definition, while returning 3-4\% on every translated split.
Moreover, the corpus was translated in two independent runs over disjoint data (verified by content hashing), and the per-check rates agree across both halves to within ${\sim}0.01$ percentage points (e.g., for French, 3.176\% vs.\ 3.179\% instruction leaks), indicating that they reflect stable properties of the translation process.
 
\begin{table}[t]
\centering
\small
\setlength{\tabcolsep}{4pt}
\begin{tabular}{lrr}
\toprule
\textbf{Measurement} & \textbf{EN source} & \textbf{Translations} \\
\midrule
Samples passing all checks & 97.2\% & 76.6--77.2\% \\
Samples repaired & 0.4\% & 17.2\% \\
Leaked instruction block & 0\% & 3.1--4.4\% \\
Upper-cased prose & 0.005\% & 0.22--5.42\% \\
\bottomrule
\end{tabular}
\caption{Rates on the untranslated English control vs.\ the translated splits under identical checks.}
\label{tab:post-control}
\end{table}
 
\subsection{Cross-Language Alignment and De-duplication}
\label{app:post-alignment}
Filtering each language independently breaks parallelism, since each language removes a slightly different subset.
We restore it by intersecting the surviving sample IDs across all five languages and rewriting every split in a shared order, recorded in a sidecar file so the correspondence can be checked externally.
The source corpus additionally contains exact-duplicate samples under distinct IDs; because translation is generative, these are textually different in every target language, so de-duplicating each language independently would break the alignment again.
We therefore de-duplicate once on the English side (hashing the normalized conversation and keeping one deterministic survivor per group) and apply the resulting drop list to all languages by ID, after the intersection.
A final verification pass streams all splits in parallel and confirms equal counts and identical IDs at every position.

\section{Data Quality Assurance via Re-annotation}
\label{app:data-quality}

The filtering pipeline in Appendix~\ref{app:dataset_filtering} operates on Propella-1 annotations of the \emph{English} source documents, which we take to transfer to the parallel translations (Section~\ref{sec:meth}).
To test this assumption directly, we run an independent, full-scale re-annotation of every translated split (DE, FR, IT, ES) with Propella-1, the same annotator used for the source; it supports more than 57 languages.
Because Propella-1 scores semantic and structural content properties (reasoning type, information density, educational value, safety) rather than surface fluency, agreement between the source and the re-annotated translations is evidence that translation preserved the properties that governed each sample's inclusion.

Table~\ref{tab:qa-de} reports the German split (${\sim}2.5$M samples) as a representative example.
The distributions concentrate on the high-quality end of every property group: the two groups with a direct source comparator (Table~\ref{tab:multilingual_dataset_stats}) track it closely (information density 94\% vs.\ 97.4\% source; educational value 88\% vs.\ 89.1\% source), and the mandatory safety/integrity labels, enforced on the source by construction (Appendix~\ref{app:dataset_filtering}), are retained above 98.6\% after translation, indicating negligible degradation.

The other three target languages follow the same pattern: good-or-excellent content quality within 98.2-98.7\%, dense-or-adequate information density within 94-96\%, and analytical-or-explanatory reasoning above 94\% in every language, with educational value and safety/integrity likewise uniform across splits.
This uniformity indicates that quality is a property of the corpus as a whole rather than of any single language.

\begin{table}[h]
\centering
\small
\begin{tabularx}{\linewidth}{@{} l X r @{}}
\toprule
\textbf{Property} & \textbf{Aggregated category} & \textbf{\%} \\
\midrule
Content quality       & good or excellent          & 98.0 \\
                      & poor or unacceptable       & 0.6 \\
\addlinespace[0.2em]
Information density   & dense or adequate          & 94.0 \\
\addlinespace[0.2em]
Educational value     & high or moderate           & 88.0 \\
\addlinespace[0.2em]
Reasoning orientation & analytical or explanatory  & 96.0 \\
                      & basic or minimal           & $<4$ \\
\addlinespace[0.2em]
Safety \& integrity   & \texttt{safe}              & $>98.6$ \\
                      & \texttt{complete\_content} & $>98.6$ \\
                      & \texttt{no\_pii}           & $>98.6$ \\
\bottomrule
\end{tabularx}
\caption{Property distribution of the re-annotated German split
(${\sim}2.5$M samples). Percentages are over annotated documents in the split.}
\label{tab:qa-de}
\end{table}

The re-annotation evidence is convergent with the downstream results in the main text.
Fine-tuning on the English split improves a strong baseline (Table~\ref{tab:multilingual_dataset_stats}, bottom-right), and fine-tuning on the translated splits induces fluent target-language reasoning whose accuracy GRPO recovers largely to, and in German and Italian beyond, base performance (Table~\ref{tab:multilingual-results-domain}); neither would be expected if the supervision signal were systematically corrupted by translation.
Successful adaptation is thus itself a form of validation of the translated traces, complementary to the annotation-level check above.
We release the per-split annotation distributions together with a high-confidence filtered subset.
As noted in the Limitations, a full native-speaker fidelity evaluation across all five languages remains future work; we regard the present annotation-level and downstream evidence as a first, convergent validation of corpus quality.

\section{Training Details}
\label{app:training-details}

\subsection{SFT Hyperparameters}
\label{app:sft-hyperparams}

Table~\ref{tab:sft-hyperparams} summarizes the hyperparameters used for supervised fine-tuning. 
The same configuration is used for all four target languages and both base models; only the dataset split differs.

\begin{table}[h]
\centering
\small
\begin{tabular}{ll}
\toprule
\textbf{Hyperparameter} & \textbf{Value} \\
\midrule
% Base model & \texttt{SmolLM3-3B} \\
Epochs & 2 \\
Max sequence length & 16{,}384 \\
Packing & Enabled \\
Precision & bfloat16 \\
Optimizer & \texttt{adamw\_torch\_fused} \\
Per-device batch size & 4 \\
Gradient accumulation steps & 4 \\
Weight decay & 0.0 \\
LR scheduler & Cosine with min LR \\
Learning Rate & $8 \times 10^{-5}$ \\
Minimum LR & $8 \times 10^{-6}$ \\
Warmup ratio & 0.1 \\
Gradient checkpointing & Enabled \\
Distributed strategy & FSDP (16 GPUs) \\
Loss & Completion-only \\
Liger kernel & Enabled \\
\bottomrule
\end{tabular}
\caption{SFT training hyperparameters (shared across all languages).}
\label{tab:sft-hyperparams}
\end{table}

\subsection{GRPO Training Details}
\label{app:grpo-details}

\paragraph{Hyperparameters}
Table~\ref{tab:grpo-hyperparams} lists the RL training configuration.
We follow the Dr.\ GRPO variant \citep{liu2025understandingr1zeroliketrainingcritical}, which removes the KL penalty and uses a normalized advantage formulation.

\begin{table}[h]
\centering
\small
\resizebox{\columnwidth}{!}{
\begin{tabular}{ll}
\toprule
\textbf{Hyperparameter} & \textbf{Value} \\
\midrule
Loss type & Dr.\ GRPO \\
Epochs & 1 \\
Generations per prompt ($G$) & 8 \\
Max completion length & 8{,}192 \\
Sampling temperature & 1.0 \\
Top-$p$ & 1.0 \\
Learning rate & $1 \times 10^{-6}$ \\
LR scheduler & Constant \\
Optimizer & AdamW ($\beta_1{=}0.9$, $\beta_2{=}0.95$) \\
Max gradient norm & 1.0 \\
Clipping $\epsilon$ & 0.2 \\
KL coefficient $\beta$ & 0.0 \\
Inner epochs & 1 \\
Per-device batch size & 4 \\
Gradient accumulation steps & 8 \\
Precision & bfloat16 \\
Distributed strategy & FSDP (8 GPUs) \\
Generation backend & vLLM (colocate mode) \\
\bottomrule
\end{tabular}
}
\caption{Dr.\ GRPO training hyperparameters.}
\label{tab:grpo-hyperparams}
\end{table}

\paragraph{Reward Function Details}

The composite reward is a weighted sum of individual components. Table~\ref{tab:reward-weights} shows the per-language reward configuration.

\begin{table}[h]
\centering
\small
\begin{tabular}{lcccc}
\toprule
\textbf{Reward Component} & \textbf{DE} & \textbf{IT} & \textbf{ES} & \textbf{FR} \\
\midrule
Accuracy & 1.0 & 1.0 & 1.0 & 1.0 \\
Language & 0.2 & 0.2 & 0.2 & 0.2 \\
Format & 0.1 & 0.1 & 0.1 & 0.1 \\
Repetition & 0.3 & 0.3 & 0.3 & 0.3 \\
\bottomrule
\end{tabular}
\caption{Reward weights per language.}
\label{tab:reward-weights}
\end{table}

\paragraph{Accuracy Reward}
The predicted answer is extracted from the last \verb|\boxed{}| expression in the completion.
Verification is attempted first via symbolic equivalence using \texttt{math\_verify}; if this fails, an exact string match is used as fallback.
The reward is binary: $1.0$ for a correct answer, $0.0$ otherwise.

\paragraph{Language Reward}
We use a fastText language identification model (\texttt{lid.176.ftz}) to score language fidelity.
The reward is computed separately over the reasoning trace (content within \texttt{<think>} tags) and the non-reasoning output (remaining text, excluding \verb|\boxed{}| expressions), then combined as $0.6 \cdot s_{\text{think}} + 0.4 \cdot s_{\text{output}}$, weighting the reasoning trace more heavily since it constitutes the majority of the generated text.

\paragraph{Format Reward}
A structured score rewards correct formatting: $+0.1$ for an opening \texttt{<think>} tag, $+0.3$ for a properly closed \texttt{<think>}...\texttt{</think>} block, $+0.1$ for the presence of a \verb|\boxed{}| answer, and a $+0.5$ bonus when the reasoning block ends before the boxed answer appears, encouraging the expected think-then-answer structure.

\paragraph{Repetition Penalty}
This component penalizes three types of degenerate repetition: (i)~\emph{consecutive $n$-gram loops} ($n \in \{1,\ldots,5\}$), detected greedily from largest to smallest $n$ to avoid double-counting; (ii)~\emph{token flooding}, penalizing tokens exceeding 15\% frequency with quadratic scaling; and (iii)~\emph{character-level repetition} of 4+ consecutive identical characters.
The raw penalty is normalized by $\sqrt{T}$ (where $T$ is the total token count) and capped at $1.0$, yielding a reward in $[-1, 0]$.

\section{Evaluation Details}
\label{app:eval-details}

All benchmarks share the same inference backend: vLLM with single-GPU tensor parallelism, bfloat16 precision, a maximum context length of 16{,}384 tokens, and 90\% GPU memory utilization.
Generation uses temperature $0.6$, top-$p$ $0.95$, and a maximum output length of 16{,}384 tokens; all benchmarks are run with seeds $\{42, 123, 456, 789, 1024\}$ (5 runs).
For final adapted models, a target-language system prompt (used during RL training) is prepended (e.g., \textit{``Du bist ein hilfreicher Assistent. Denke zuerst in \texttt{<think>}-Tags nach und gib dann deine Antwort.''} for German); base and SFT-trained models are evaluated without a system prompt.

\subsection{Science Tasks}

\paragraph{MGSM}
We use the \texttt{juletxara/mgsm} dataset, evaluating on each target-language split. %\footnote{Italian uses separately prepared splits released with the corpus, as it is not included in the original MGSM, MT-Math100, and GPQA Diamond releases.}
Answer extraction follows a three-stage fallback: (i)~the content of the last \verb|\boxed{}| expression, (ii)~the number following a \texttt{\#\#\#\#} delimiter, and (iii)~the last number in the generated text.
Extracted values are normalized by removing thousand separators (both dot and comma conventions), converting European decimal notation, and truncating trailing zeros before comparison with the integer gold answer.

\paragraph{MT-Math100}
We use the \texttt{amphora/MCLM} dataset (MT-MATH100 configuration).
The predicted answer is extracted from the last \verb|\boxed{}| expression using nested-brace-aware parsing, which correctly handles expressions such as \verb|\boxed{\frac{1}{2}}|.
Evaluation is exact string match between the extracted prediction and the gold answer.

\paragraph{GPQA Diamond (Multilingual)}
We use the \texttt{shanchen/gpqa\_diamond\_mc\_multilingual} dataset, a four-option multiple-choice benchmark (A--D).
The predicted label is extracted first from a \verb|\boxed{}| expression containing a single letter; if absent, the last standalone A--D token in the output is used.
The gold label is extracted from the solution field using the same procedure.
Accuracy is the fraction of exact label matches.

\subsection{Non-Science Tasks}

\paragraph{Global MMLU Lite}
We use the \texttt{CohereLabs/Global-MMLU-Lite} dataset, evaluating on each target-language split.
Each prompt presents the question, four labeled options (A-D), and a localized \verb|\boxed{}| instruction (e.g., \textit{``Bitte geben Sie Ihre endgültige Antwort in der Form \textbackslash boxed\{A\}, \textbackslash boxed\{B\}, \textbackslash boxed\{C\} oder \textbackslash boxed\{D\} an.''} for German).
The predicted label is extracted from the last \verb|\boxed{}| match; if absent, a fallback searches for the first occurrence of a standalone A-D letter in the output.
Per-subject accuracy is additionally logged for fine-grained analysis.

\paragraph{Belebele}
We use the \texttt{facebook/belebele} dataset with BCP-47 script-tagged splits (e.g., \texttt{deu\_Latn} for German).
Each prompt comprises a localized header (e.g., \textit{``Lesen Sie den Text und beantworten Sie die Frage.''}), a reading passage, a question, four answer options, and a \verb|\boxed{}| instruction.
Answer extraction follows the same procedure as Global MMLU Lite.
The gold label is derived from the 1-indexed \texttt{correct\_answer\_num} field mapped to A-D.

\paragraph{INCLUDE}
We use the \texttt{CohereLabs/include-lite-44} dataset, loading each language by its full name (e.g., \texttt{German}).
Prompts follow the same format as Global MMLU Lite: question, four labeled options, and a localized \verb|\boxed{}| instruction.
The gold answer is a 0-indexed integer mapped to A--D.
Answer extraction and per-subject logging follow the same procedure as Global MMLU Lite.

\section{Computational Budget}
\label{app:compute}

All translation and experiments were conducted on NVIDIA B200 GPUs.
Corpus translation with \texttt{Qwen3-32B} consumed approximately 15{,}550 GPU hours across the four target languages.
Each SFT run used 16 GPUs via FSDP: the 3B runs took approximately 20 hours of wall-clock time each (1{,}600 GPU hours over the four target languages and the English baseline), and the 7B runs approximately 30 hours each (1{,}920 GPU hours over the four target languages).
Each GRPO run used a single 8-GPU node: approximately 21 hours per 3B run (840 GPU hours in total) and 29 hours per 7B run (928 GPU hours in total).
% VERIFY: 840 implies 5 GRPO runs at 21h x 8 GPUs -- was there an English GRPO run? If only 4 runs, the wall-clock or total needs correcting.
In total, adaptation of both base models consumed approximately 5{,}288 GPU hours, bringing the full compute footprint, including translation, to approximately 20{,}838 GPU hours, excluding data filtering and evaluation.
% VERIFY: or add filtering/evaluation hours and update the total.

\section{Artifacts and Licenses}
\label{app:licenses}

Table~\ref{tab:licenses} lists all artifacts used in this work together with their licenses.
All source datasets, models, and evaluation benchmarks are publicly released for research use, and their licenses permit the derivative use made here: sampling and translating source corpora into \reasonxl, fine-tuning the base models, and evaluating on the listed benchmarks.
\reasonxl\ will be released under CC-BY-4.0% TODO: pick your release license; must be compatible with the most restrictive source license below
, with metadata identifying the source dataset of every sample.

\begin{table}[h]
\centering
\small
\setlength{\tabcolsep}{4pt}
\begin{tabular}{ll}
\toprule
\textbf{Artifact} & \textbf{License} \\
\midrule
\multicolumn{2}{l}{\textit{Source datasets}} \\
Nemotron-Cascade-SFT & CC-BY-4.0 \\
Dolci-Think-SFT (Olmo~3) & ODC-BY \\
Llama-Nemotron-PTD & CC-BY-4.0 \\
Nemotron-Science-v1 & CC-BY-4.0 \\
Nemotron-IF-Chat-v1 & CC-BY-4.0 \\
\addlinespace[0.2em]
\multicolumn{2}{l}{\textit{Models}} \\
\texttt{Qwen3-32B} (translation) & Apache-2.0 \\
\texttt{Propella-1-4B} (annotation) & Apache-2.0 \\
\texttt{SmolLM3-3B} (base) & Apache-2.0 \\
\texttt{OpenThinker3-7B} (base) & Apache-2.0 \\
\addlinespace[0.2em]
\multicolumn{2}{l}{\textit{Training / evaluation data}} \\
MATH, GSM8K (RL prompts) & MIT \\
MGSM & CC-BY-SA-4.0 \\
MT-Math100 (MCLM) & CC-BY-4.0 \\
GPQA Diamond (multiling.) & CC-BY-4.0 \\
Global MMLU Lite & Apache-2.0 \\
Belebele & CC-BY-SA-4.0 \\
INCLUDE & CC-BY-4.0 \\
\bottomrule
\end{tabular}
\caption{Licenses of artifacts used in this work.
}
\label{tab:licenses}
\end{table}

\section{Detailed Per-Benchmark Results}
\label{app:detailed-results}

Tables~\ref{tab:detailed-smollm} and~\ref{tab:detailed-openthinker} report per-benchmark scores for all four target languages, corresponding to the domain averages in Table~\ref{tab:multilingual-results-domain}.
Scores are means over 5 runs with different random seeds; $\pm$ denotes standard deviation, and \textbf{bold} marks the best method per language and column.

\begin{table*}[t]
\centering
\small
\setlength{\tabcolsep}{5pt}
\begin{tabular}{llccc|ccc|c|c}
\toprule
& & \multicolumn{3}{c|}{\faFlask\ \textit{Science}} & \multicolumn{3}{c|}{\faLanguage\ \textit{Lang.}} & & \\
\cmidrule(lr){3-5} \cmidrule(lr){6-8}
\textbf{Lang} & \textbf{Method} & \textit{MATH100} & \textit{GPQA} & \textit{MGSM} & \textit{Bele.} & \textit{GMMLU} & \textit{INCL.} & \textit{Avg.} & \textit{\%TL} \\
\midrule
\rowcolor{rowgray}
\textsc{de} & SmolLM3 & $66.1{\scriptstyle \pm 1.4}$ & $\boldsymbol{36.9{\scriptstyle \pm 3.5}}$ & $\boldsymbol{81.8{\scriptstyle \pm 2.1}}$ & $81.5{\scriptstyle \pm 1.3}$ & $\boldsymbol{66.5{\scriptstyle \pm 0.7}}$ & $\boldsymbol{48.5{\scriptstyle \pm 1.3}}$ & $63.5$ & $0.0$ \\
 & \quad + SFT & $51.5{\scriptstyle \pm 2.1}$ & $30.5{\scriptstyle \pm 3.0}$ & $70.8{\scriptstyle \pm 2.3}$ & $82.7{\scriptstyle \pm 0.5}$ & $55.2{\scriptstyle \pm 1.9}$ & $44.3{\scriptstyle \pm 1.3}$ & $55.8$ & $96.5$ \\
\rowcolor{lightolivegreen-3!50}
 & \quad + GRPO & $\boldsymbol{70.5{\scriptstyle \pm 2.7}}$ & $34.0{\scriptstyle \pm 2.8}$ & $80.4{\scriptstyle \pm 1.2}$ & $\boldsymbol{85.3{\scriptstyle \pm 0.7}}$ & $66.4{\scriptstyle \pm 2.0}$ & $45.8{\scriptstyle \pm 2.6}$ & $\boldsymbol{63.7}$ & $100.0$ \\
\midrule
\rowcolor{rowgray}
\textsc{it} & SmolLM3 & $64.4{\scriptstyle \pm 0.8}$ & $\boldsymbol{39.4{\scriptstyle \pm 3.2}}$ & $82.9{\scriptstyle \pm 1.7}$ & $75.9{\scriptstyle \pm 0.7}$ & $63.7{\scriptstyle \pm 1.6}$ & $63.1{\scriptstyle \pm 1.2}$ & $64.9$ & $0.0$ \\
 & \quad + SFT & $58.6{\scriptstyle \pm 1.8}$ & $33.8{\scriptstyle \pm 1.2}$ & $82.0{\scriptstyle \pm 1.4}$ & $84.0{\scriptstyle \pm 1.3}$ & $65.5{\scriptstyle \pm 0.8}$ & $64.1{\scriptstyle \pm 3.2}$ & $64.6$ & $96.2$ \\
\rowcolor{lightolivegreen-3!50}
 & \quad + GRPO & $\boldsymbol{71.3{\scriptstyle \pm 2.9}}$ & $36.9{\scriptstyle \pm 1.2}$ & $\boldsymbol{87.2{\scriptstyle \pm 1.0}}$ & $\boldsymbol{85.8{\scriptstyle \pm 0.7}}$ & $\boldsymbol{69.5{\scriptstyle \pm 1.3}}$ & $\boldsymbol{70.4{\scriptstyle \pm 1.6}}$ & $\boldsymbol{70.2}$ & $100.0$ \\
\midrule
\rowcolor{rowgray}
\textsc{es} & SmolLM3 & $\boldsymbol{63.4{\scriptstyle \pm 1.7}}$ & $\boldsymbol{37.5{\scriptstyle \pm 3.4}}$ & $\boldsymbol{81.4{\scriptstyle \pm 1.1}}$ & $83.6{\scriptstyle \pm 0.4}$ & $\boldsymbol{68.4{\scriptstyle \pm 1.4}}$ & $61.8{\scriptstyle \pm 1.1}$ & $\boldsymbol{66.0}$ & $0.0$ \\
 & \quad + SFT & $57.4{\scriptstyle \pm 2.1}$ & $30.9{\scriptstyle \pm 0.8}$ & $79.6{\scriptstyle \pm 1.8}$ & $84.5{\scriptstyle \pm 0.8}$ & $64.3{\scriptstyle \pm 1.1}$ & $58.5{\scriptstyle \pm 2.9}$ & $62.5$ & $78.3$ \\
\rowcolor{lightolivegreen-3!50}
 & \quad + GRPO & $53.5{\scriptstyle \pm 2.3}$ & $34.4{\scriptstyle \pm 2.4}$ & $74.4{\scriptstyle \pm 1.8}$ & $\boldsymbol{84.8{\scriptstyle \pm 0.8}}$ & $67.6{\scriptstyle \pm 1.3}$ & $\boldsymbol{63.5{\scriptstyle \pm 1.5}}$ & $63.0$ & $100.0$ \\
\midrule
\rowcolor{rowgray}
\textsc{fr} & SmolLM3 & $\boldsymbol{65.3{\scriptstyle \pm 1.0}}$ & $\boldsymbol{36.1{\scriptstyle \pm 1.3}}$ & $\boldsymbol{77.7{\scriptstyle \pm 2.4}}$ & $85.5{\scriptstyle \pm 0.8}$ & $\boldsymbol{65.9{\scriptstyle \pm 1.9}}$ & $\boldsymbol{56.6{\scriptstyle \pm 1.4}}$ & $\boldsymbol{64.5}$ & $0.0$ \\
 & \quad + SFT & $58.8{\scriptstyle \pm 3.0}$ & $30.7{\scriptstyle \pm 2.6}$ & $76.1{\scriptstyle \pm 3.3}$ & $85.9{\scriptstyle \pm 1.1}$ & $63.6{\scriptstyle \pm 1.5}$ & $50.4{\scriptstyle \pm 2.1}$ & $60.9$ & $95.8$ \\
\rowcolor{lightolivegreen-3!50}
 & \quad + GRPO & $63.0{\scriptstyle \pm 0.8}$ & $33.8{\scriptstyle \pm 3.2}$ & $69.8{\scriptstyle \pm 2.4}$ & $\boldsymbol{86.3{\scriptstyle \pm 0.8}}$ & $63.5{\scriptstyle \pm 1.9}$ & $54.6{\scriptstyle \pm 0.9}$ & $61.8$ & $100.0$ \\
\bottomrule
\end{tabular}
\caption{Performance on downstream target-language tasks for \textsc{SmolLM3-3B}.} % Every score is an average over five runs.}
\label{tab:detailed-smollm}
\end{table*}

\begin{table*}[t]
\centering
\small
\setlength{\tabcolsep}{5pt}
\begin{tabular}{llccc|ccc|c|c}
\toprule
& & \multicolumn{3}{c|}{\faFlask\ \textit{Science}} & \multicolumn{3}{c|}{\faLanguage\ \textit{Lang.}} & & \\
\cmidrule(lr){3-5} \cmidrule(lr){6-8}
\textbf{Lang} & \textbf{Method} & \textit{MATH100} & \textit{GPQA} & \textit{MGSM} & \textit{Bele.} & \textit{GMMLU} & \textit{INCL.} & \textit{Avg.} & \textit{\%TL} \\
\midrule
\rowcolor{rowgray}
\textsc{de} & OpenThinker3-7B & $\boldsymbol{69.7{\scriptstyle \pm 1.3}}$ & $\boldsymbol{44.3{\scriptstyle \pm 1.0}}$ & $70.0{\scriptstyle \pm 1.2}$ & $78.1{\scriptstyle \pm 1.0}$ & $59.3{\scriptstyle \pm 0.9}$ & $37.8{\scriptstyle \pm 3.1}$ & $59.9$ & $0$ \\
 & \quad + SFT & $25.1{\scriptstyle \pm 2.8}$ & $25.4{\scriptstyle \pm 4.1}$ & $50.6{\scriptstyle \pm 2.6}$ & $62.5{\scriptstyle \pm 1.4}$ & $47.4{\scriptstyle \pm 1.6}$ & $44.9{\scriptstyle \pm 2.4}$ & $42.6$ & $87.3$ \\
\rowcolor{lightolivegreen-3!50}
 & \quad + GRPO & $63.0{\scriptstyle \pm 2.0}$ & $28.6{\scriptstyle \pm 1.5}$ & $\boldsymbol{81.4{\scriptstyle \pm 0.8}}$ & $\boldsymbol{78.6{\scriptstyle \pm 0.9}}$ & $\boldsymbol{61.0{\scriptstyle \pm 0.9}}$ & $\boldsymbol{51.7{\scriptstyle \pm 1.7}}$ & $\boldsymbol{60.7}$ & $100.0$ \\
\midrule
\rowcolor{rowgray}
\textsc{it} & OpenThinker3-7B & $\boldsymbol{67.5{\scriptstyle \pm 1.6}}$ & $\boldsymbol{42.2{\scriptstyle \pm 2.1}}$ & $74.2{\scriptstyle \pm 0.6}$ & $79.0{\scriptstyle \pm 1.1}$ & $59.3{\scriptstyle \pm 2.5}$ & $57.8{\scriptstyle \pm 2.6}$ & $63.3$ & $0$ \\
 & \quad + SFT & $27.5{\scriptstyle \pm 4.0}$ & $24.7{\scriptstyle \pm 1.7}$ & $55.0{\scriptstyle \pm 2.3}$ & $65.8{\scriptstyle \pm 1.5}$ & $47.0{\scriptstyle \pm 2.5}$ & $45.2{\scriptstyle \pm 2.6}$ & $44.2$ & $92.0$ \\
\rowcolor{lightolivegreen-3!50}
 & \quad + GRPO & $61.4{\scriptstyle \pm 2.2}$ & $29.1{\scriptstyle \pm 3.9}$ & $\boldsymbol{84.0{\scriptstyle \pm 2.0}}$ & $\boldsymbol{82.6{\scriptstyle \pm 0.6}}$ & $\boldsymbol{65.1{\scriptstyle \pm 1.4}}$ & $\boldsymbol{66.1{\scriptstyle \pm 2.0}}$ & $\boldsymbol{64.7}$ & $100.0$ \\
\midrule
\rowcolor{rowgray}
\textsc{es} & OpenThinker3-7B & $\boldsymbol{67.9{\scriptstyle \pm 1.6}}$ & $\boldsymbol{49.7{\scriptstyle \pm 3.7}}$ & $\boldsymbol{89.9{\scriptstyle \pm 0.8}}$ & $\boldsymbol{85.8{\scriptstyle \pm 0.5}}$ & $\boldsymbol{67.0{\scriptstyle \pm 1.5}}$ & $\boldsymbol{59.3{\scriptstyle \pm 1.4}}$ & $\boldsymbol{69.9}$ & $0$ \\
 & \quad + SFT & $34.9{\scriptstyle \pm 2.6}$ & $27.6{\scriptstyle \pm 1.2}$ & $63.4{\scriptstyle \pm 2.2}$ & $72.4{\scriptstyle \pm 2.1}$ & $56.3{\scriptstyle \pm 1.7}$ & $55.4{\scriptstyle \pm 2.4}$ & $51.7$ & $72.5$ \\
\rowcolor{lightolivegreen-3!50}
 & \quad + GRPO & $49.5{\scriptstyle \pm 1.7}$ & $31.3{\scriptstyle \pm 1.4}$ & $64.4{\scriptstyle \pm 0.9}$ & $78.6{\scriptstyle \pm 0.6}$ & $63.7{\scriptstyle \pm 1.7}$ & $57.9{\scriptstyle \pm 1.2}$ & $57.6$ & $100.0$ \\
\midrule
\rowcolor{rowgray}
\textsc{fr} & OpenThinker3-7B & $\boldsymbol{69.9{\scriptstyle \pm 0.8}}$ & $\boldsymbol{47.9{\scriptstyle \pm 2.6}}$ & $\boldsymbol{84.0{\scriptstyle \pm 1.2}}$ & $\boldsymbol{87.9{\scriptstyle \pm 0.5}}$ & $\boldsymbol{65.1{\scriptstyle \pm 1.5}}$ & $52.3{\scriptstyle \pm 0.5}$ & $\boldsymbol{67.9}$ & $0$ \\
 & \quad + SFT & $26.3{\scriptstyle \pm 1.3}$ & $28.3{\scriptstyle \pm 3.0}$ & $46.4{\scriptstyle \pm 2.5}$ & $74.2{\scriptstyle \pm 0.7}$ & $51.3{\scriptstyle \pm 2.1}$ & $45.9{\scriptstyle \pm 3.3}$ & $45.4$ & $80.9$ \\
\rowcolor{lightolivegreen-3!50}
 & \quad + GRPO & $52.5{\scriptstyle \pm 3.4}$ & $31.6{\scriptstyle \pm 3.5}$ & $71.8{\scriptstyle \pm 1.7}$ & $81.9{\scriptstyle \pm 1.0}$ & $60.2{\scriptstyle \pm 1.9}$ & $\boldsymbol{54.0{\scriptstyle \pm 1.4}}$ & $58.7$ & $100.0$ \\
\bottomrule
\end{tabular}
\caption{Performance on downstream target-language tasks for \textsc{OpenThinker3-7B}.} % Every score is an average over five runs.}
\label{tab:detailed-openthinker}
\end{table*}

\end{document}